\documentclass[runningheads]{llncs}
\usepackage[T1]{fontenc}
\usepackage{graphicx}
\usepackage{amssymb}
\usepackage[misc]{ifsym}

\begin{document}
\title{HDCompression: Hybrid-Diffusion Image Compression for Ultra-Low Bitrates
}
\titlerunning{Hybrid-Diffusion Image Compression for Ultra-Low Bitrates}
%
\author{Lei Lu\inst{1*} \and Yize Li\inst{1*} \and Yanzhi Wang\inst{1} \and Wei Wang\inst{2} \and Wei Jiang\inst{2} \textsuperscript{\Letter}
}

\authorrunning{L. Lu et al.}
%
\institute{Northeastern University, Boston MA 02115, USA \and
Futurewei Technologies Inc., San Jose CA 95131, USA\\
\email{\{lu.le1, li.yize, yanz.wang\}@northeastern.edu, \\
\{rickweiwang, wjiang\}@futurewei.com}}
\maketitle              
\footnotetext{*Both authors contributed equally.}

\begin{abstract}
Image compression under ultra-low bitrates remains challenging for both conventional learned image compression (LIC) and generative vector-quantized (VQ) modeling. Conventional LIC suffers from severe artifacts due to heavy quantization, while generative VQ modeling gives poor fidelity due to the mismatch between learned generative priors and specific inputs. In this work, we propose Hybrid-Diffusion Image Compression (HDCompression), a dual-stream framework that utilizes both generative VQ-modeling and diffusion models, as well as conventional LIC, to achieve both high fidelity and high perceptual quality. Different from previous hybrid methods that directly use pre-trained LIC models to generate low-quality fidelity-preserving information from heavily quantized latent, we use diffusion models to extract high-quality complementary fidelity information from the ground-truth input, which can enhance the system performance in several aspects: improving index map prediction, enhancing the fidelity-preserving output of the LIC stream, and refining conditioned image reconstruction with VQ-latent correction. In addition, our diffusion model is based on a dense representative vector (DRV), which is lightweight with very simple sampling schedulers. Extensive experiments demonstrate that our HDCompression outperforms the previous conventional LIC, generative VQ-modeling, and hybrid frameworks in both quantitative metrics and qualitative visualization, providing balanced robust compression performance at ultra-low bitrates.

\end{abstract}

\section{Introduction}
\label{sec:intro}
\begin{figure}
\centering
  \includegraphics[width=\textwidth]     {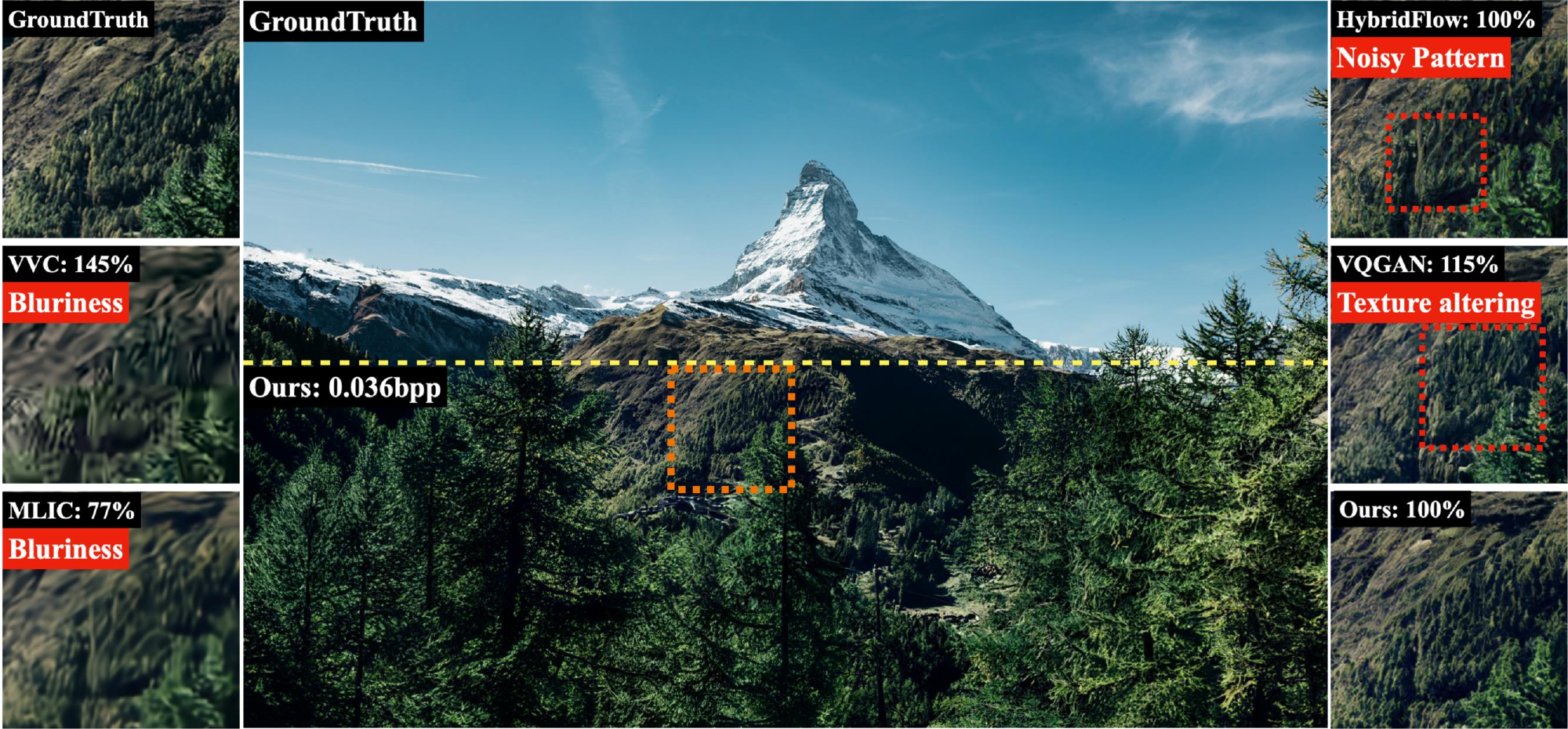}
    \caption{Visual comparisons of different methods. Bitrates are listed as percentages relative to our method. Traditional hand-crafted VVC and conventional LIC method MLIC present severe blurs, single-streamed VQ-codebook-based VQGAN generates inauthentic details, and HybridFlow has high-frequency artifacts. Our HDCompression retains both fidelity and clarity.}
    \label{fig:general_compare}
  \label{fig:teaser}
\end{figure}

There has been an explosion of applications requiring transmitting large amounts of image data with limited bandwidth, calling for effective image compression solutions at ultra-low bitrates. Despite decades of research~\cite{jpegai,BPG,vvc,JPEG2000}, image compression at ultra-low bitrates remains an ongoing challenge. This is primarily due to the conventional framework of applying heavy quantization at ultra-low bitrates, resulting in significant artifacts. In the conventional framework, an encoder first transforms the input image into a latent feature, either by traditional transformation as in JPEG~\cite{JPEG2000} or by neural network models as in learned image compression (LIC)~\cite{jpegai}. Then the latent feature is quantized by rounding operations for transmission. A decoder subsequently recovers the output image from the dequantized latent feature using traditional inverse-transformation or neural network models. Therefore, bit reduction occurs during the quantization process. Especially at ultra-low bitrates, intense quantization causes excessive information loss, leading to severe and unpleasant blurriness and noises. Although numerous efforts~\cite{jpegai,losslesshyperprior,mlic2023} have been focused on improving the transformation model and prediction of quantization statistics, such artifacts cannot be easily mitigated, as shown in Fig.~\ref{fig:general_compare}. 

Besides Variational AudoEncoder (VAE), generative methods such as generative adversarial networks (GANs)~\cite{goodfellow2014generative,isola2017image,mirza2014conditional,radford2015unsupervised} and diffusion models~\cite{dhariwal2021diffusion,ho2020denoising,kong2025,li2024pruning,rombach2022high,zeng2025enhancing} offer promising opportunities to explore alternative frameworks for image compression. Generative approaches learn statistical priors from images, which allows for the synthesis of perceptually realistic high-quality image details from degraded input images. For instance, vector-quantized (VQ) image modeling~\cite{esser2021taming,van2017neural} has been recently used for image compression~\cite{jia2024generative,WACV2024,vqganmasiwei}, where the learned generative priors serve as visual codewords that span a latent space, enabling images to be mapped into vector-quantized integer indices. 
Thus, the learned VQ latent space provides a refined quantization strategy that retrieves high-quality, information-rich codewords for reconstructing high-realism outputs, which could lead to finer quantization adjustments and effectively avoid the degraded outputs at ultra-low bitrates. However, while the generated outputs are visually appealing to human eyes, the learned generative priors (\textit{i.e.}, codewords) often deviate from authentic image details, bringing about significant pixel-level differences from the original inputs. Thus, most VQ-modeling-based approaches~\cite{jia2024generative,vqganmasiwei} primarily address perceptual quality only and operate at extremely low bitrates where poor fidelity may be tolerated, as shown in Fig.~\ref{fig:general_compare}. 

For the practical task of image compression, both content authenticity and visual quality are crucial, even at ultra-low bitrates. However, there is a complex and contradictory relationship between perceptual quality and fidelity~\cite{blau2018perception}, making it very challenging for a method to perfect both aspects for general scenarios. Recently, HybridFlow~\cite{lu2024hybridflow} has synergized the conventional LIC and the generative VQ-modeling to preserve both fidelity and perceptual quality at ultra-low bitrates (around 0.05 bpp). In HybridFlow, VQ-modeling provides high-realism generation, while conventional LIC offers authentic details from each specific input. However, after incorporating conventional LIC directly into HybridFlow, the quantization issue at ultra-low bitrates still severely impacts the assistive fidelity information quality for indices map prediction and conditional reconstruction. 

To address the issues above and maintain the balance between fidelity and perceptual quality, we propose the Hybrid-Diffusion Compression (HDCompression) approach that effectively exploits both GAN and diffusion models (DM) for ultra-low-bitrate image compression in a dual manner. The diffusion model contributes detailed perceptual features that address high-fidelity requirements, while the VQ-based stream provides structured latent codebooks that enable efficient compression at ultra-low bitrates. Furthermore, instead of directly using pre-trained DM as previous DM-based compression methods~\cite{relic2024lossy}, we utilize a dense representative vector (DRV) to mitigate the heavy computation and memory consumption issues. Compared with previous state-of-the-art (SOTA) image compression methods, including traditional VVC, single-stream LIC method MLIC~\cite{mlic2023}, single-stream VQ-modeling method~\cite{vqganmasiwei}, and dual-stream HybridFlow~\cite{lu2024hybridflow}, HDCompression can improve perceptual quality (LPIPS) by 26\% over HybridFlow, while maintaining the same level of fidelity (PSNR), and can reduce artifacts like random or structured noise patterns.  
We highlight our contributions as follows.
\begin{itemize}
    \item We introduce a novel dual-stream framework, HDCompression. The generative stream exploits the power of both VQ-based modeling and diffusion-based latent structure learning, where codebooks provide general image priors for reconstruction and the lightweight diffusion module learns structural priors of joint embedding between high-quality original inputs and low-quality compressed inputs. Based on the compressed inputs, the diffusion module recovers the DRV in decoder to provide input-specific fidelity information without additional transmission, which complements the fidelity information of the conventional LIC stream to enhance the performance of both indices recovery and final reconstruction.    
    \item We design an efficient DRV-based lightweight vector-wise diffusion module with a 4-step sampling scheduler to provide complementary fidelity information instead of modeling entire image structures. This approach mitigates the difficulty of obtaining accurate denoising guidance at ultra-low bitrates and reduces computation and memory requirements. 
    \item We propose two modules to merge the generative stream and the conventional LIC stream. The enhancement module uses the DRV from the generative stream to improve the conventional LIC stream, providing improved fidelity information for the reconstruction. The VQ-codebook latent correction module uses the enhanced conventional LIC stream to reduce VQ loss during indices prediction. 
\end{itemize}
\begin{figure*}[htbp]
    \includegraphics[width=\linewidth]
    {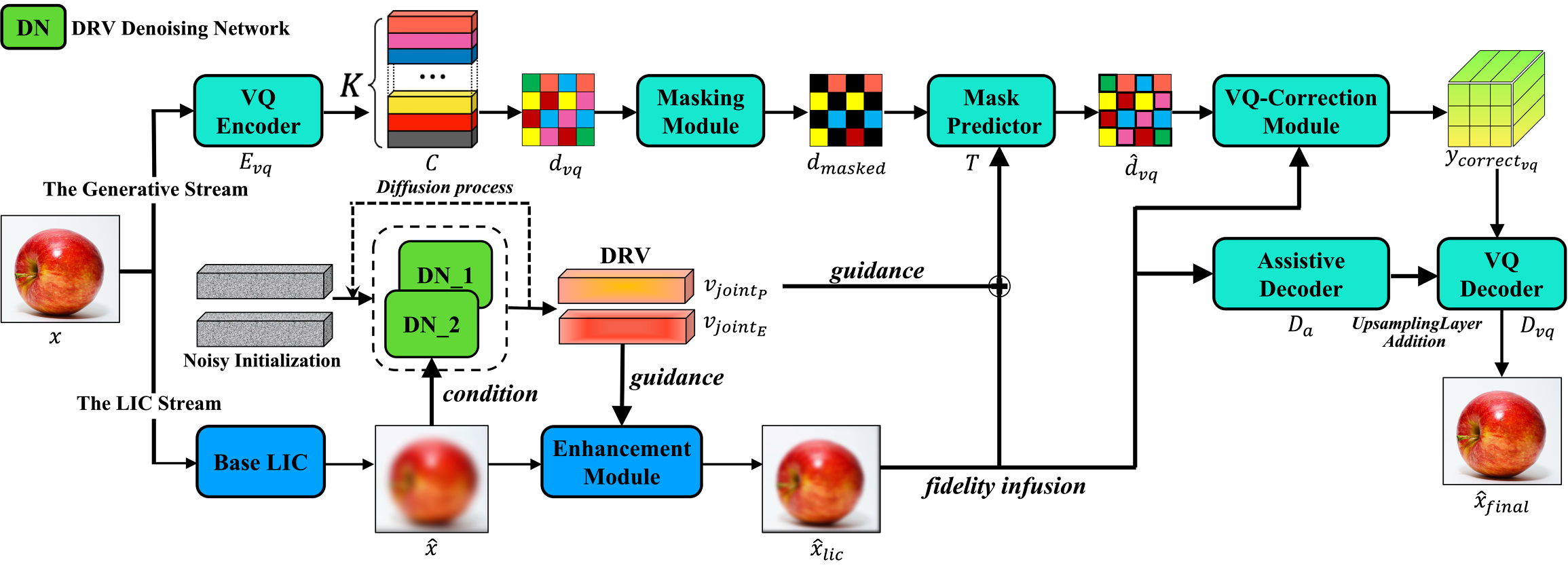}
    \caption{System Overview. We sample 2 Dense Representative Vectors (\textbf{DRVs}) by Denoising Networks (\textbf{DNs}) conditioned on the base LIC output $\hat{\textbf{x}}$. These DRVs serve as global guidance for enhancing fidelity and mask prediction. The enhanced LIC output $\hat{\textbf{x}}_{lic}$ further infuses fidelity information into the mask predictor and VQ Decoder in the generative stream.}
    \label{fig:system_overview}
\end{figure*}

\section{Related Work}
\label{sec:relate_work}
\subsection{Learned Image Compression} LIC~\cite{jpegai,Cheng2020,mlic2023} using neural networks~\cite{Hyperprior2018,li2024nas,ma2019image} has shown superior performance over traditional methods like JPEG~\cite{JPEG2000}, VVC~\cite{vvc}. One most popular LIC frameworks is based on VAE. In the encoder, the input image is encoded into a dense latent feature, which is quantized by rounding operations for efficient transmission. Then the decoder reconstructs the output image based on the dequantized latent feature. At ultra-low bitrates, the information loss caused by the universal rounding quantization is too damaging to be recovered by the decoder, resulting in severe artifacts like blurriness, noises, blocky effects, \textit{etc.}

\subsection{Image Compression by Generative Models}
Generative models such as GAN and diffusion models have been used for image compression in recent years. 

\noindent \textbf{VQ-codebook-based image compression.} 
GAN-assisted VQ-codebook methods have shown advantages over traditional LICs, especially in perceptual quality~\cite{WACV2024,vqganmasiwei,muckley2023improving}. A discrete codebook maps the encoded image latent into a transmitted integer indices map instead of the conventional rounding-quantized latent, enabling lower bitrate and greater resilience to network fluctuations. The decoder retrieves the vector-quantized (VQ) latent from the shared codebook according to the indices map and then reconstructs the image. To improve the visual quality and fidelity of the reconstruction from the VQ-based generator, the GAN discriminator is either applied on the pixel level~\cite{WACV2024,vqganmasiwei} or further extended to the indices map~\cite{muckley2023improving}. However, the VQ-codebook captures general image priors and often deviates from individual image details. As a result, although visually appealing, the reconstructed images usually present significant pixel-level distortions (\textit{e.g.}, poor PSNR).

\noindent \textbf{Image compression with diffusion models.}
DMs have surpassed GAN in many vision tasks. Based on VAE and latent diffusion, latent diffusion models (LDMs) can be directly applied to image compression with minor changes. For example, hyperpriors extracted from the input are used as conditional information for the multi-step denoising process to recover a denoised latent for reconstruction~\cite{yang2024lossy}. The performance suffers at ultra-low bitrates as the noisy initialization requires relatively accurate denoising guidance. Also, many steps ($>$15) are usually required, causing severe computation latency. The problem may be alleviated by exploiting the strong generative ability of pre-trained LDM like Stable Diffusion (SD), where the rounding-quantized latent is refined by inverse diffusion~\cite{relic2024lossy} with adaptive denoising steps. However, the severe memory burden of pre-trained SD models (often with $>$1.3B parameters) is impractical for compression.  

\noindent \textbf{Hybrid dual-stream image compression.}
A dual-stream LIC framework has been proposed recently, 
which combines the VQ-codebook-based compression and conventional LIC and takes advantage of the resulting synergy at ultra-low bitrates. For example, HybridFlow~\cite{lu2024hybridflow} uses pre-trained conventional LIC models to provide fidelity information, which assists the VQ-codebook-based stream in both indices prediction and generative reconstruction. The dual-stream framework aims to achieve a balanced reconstruction quality between fidelity and perception at ultra-low bitrates. However, conventional LIC methods are directly equipped into the entire system without any adaptation. The quality of the assisting fidelity information still suffers from the common rounding quantization issue of general LIC. Inspired by the dual-stream framework, we use DMs in this paper to effectively provide complementary input-specific fidelity information to boost performance further.

\subsection{Dense-Vector-based Vision Model} For convolution-based vision models~\cite{cavigelli2017cas,li2023less,li2025achieving,li2025frequency}, an input image is commonly encoded into a latent feature $L$ as a 3D tensor~\cite{gong2022reverse,li2025sepprune,li2024sglp,yang2025fairsmoe}. As transformer blocks gain popularity in vision models, it has been shown that a highly dense 1D vector $V$ is quite powerful to serve as conditional guidance for various downstream tasks. In general, $V$ carries input-adaptive information learned for specific tasks, which conditionally modify $L$ to improve performance over individual inputs. For instance, RCG~\cite{RCG2023} uses a global guidance $V$ as a condition for image generation. DiffIR~\cite{xia2023diffir} uses a joint embedding $V$ between the ground-truth and degraded inputs to provide ground-truth information for guiding image restoration. In this work, we incorporate such a dense vector $V$ to provide complementary input-specific fidelity information for improved reconstruction.
\section{Methodology}
\label{sec:method}
As shown in Fig.~\ref{fig:system_overview}, the proposed HDCompression approach has two main data streams: a generative stream and a conventional LIC stream.  

\subsection{The LIC Stream}

For general LIC, an input image $\textbf{x}\in\mathbb{R}^{H\!\times\!W\! \times\!3}$ is first encoded into a latent $\textbf{y}\in\mathbb{R}^{\frac{H}{m}\!\times\!\frac{W}{m}\!\times\!c}$ by an LIC encoder. Then $\textbf{y}$ is rounding quantized into $\textbf{y}_{q}$ for easy transmission, and a LIC decoder reconstructs the output image ${\hat{\textbf{x}}}$ from received $\textbf{y}_{q}$. The downsampling factor $m$ and latent dimension $c$ determine  $\textbf{y}$'s size, \textit{e.g.}, a larger $\textbf{y}$ that has more representative capacity, gives better reconstruction but consumes more bits. 

In the dual-stream framework, the LIC stream provides the fidelity information to the final reconstruction. At ultra-low bitrates, such information quality severely suffers due to the large rounding loss. In this work, instead of merely relying on pre-trained LIC models, we introduce a DRV-diffusion-based enhancement module to improve the fidelity of information from the LIC stream. Following the DRV-based vision model, we leverage a DRV to carry ground-truth information from the current input $\textbf{x}$ to enhance the fidelity of the LIC stream. 
In detail, the structure of the DRV-based enhancement module is inspired by DiffIR~\cite{xia2023diffir}, where a DRV $\textbf{v}_{gt}$ containing ground-truth information is fused into Restormer via cross-attention in transformer blocks.  However, $\textbf{v}_{gt}$ is too heavy to transfer for ultra-low-bitrate compression. Therefore, we propose to regenerate a DRV in the decoder by utilizing a joint-embedding DRV vector $\textbf{v}_{joint_{E}}$:
\begin{equation}
    \textbf{v}_{joint_{E}} = E_{L}(\textbf{x}, \hat{\textbf{x}}),
\end{equation}
where $E_{L}$ is a DRV extractor that takes the concatenation of $\textbf{x}$ and $\hat{\textbf{x}}$ as input and generates $\textbf{v}_{joint_{E}}$ as the DRV in the joint space between $\textbf{x}$ and $\hat{\textbf{x}}$. We sample DRV $\hat{\textbf{v}}_{joint_{E}}$ in the decoder via a lightweight diffusion model that is conditioned on the decompressed output $\hat{\textbf{x}}$ of the pre-trained LIC. This better constraints the denoising process to generate the output to be consistent with the content of the original $\textbf{x}$.

Specifically, the forward diffusion process on $\textbf{v}_{joint_{E}}$ with $T$ total steps can be described as:
\begin{equation}
    \textbf{v}_{joint_{E},T} = \sqrt{\bar{\alpha}_T} \mathbf{v}_{joint_{E}} + \sqrt{1 - \bar{\alpha}_T} \mathbf{z}_T, \label{eqn:diffusion_drv1}
\end{equation}
where \(\mathbf{z}_T \sim \mathcal{N}(\mathbf{0}, \mathbf{I})\) is the Gaussian noise. \(\!\bar{\alpha}_T\) is accumulated dot product of pre-defined intensity factors $\beta$: 
\begin{equation}
    \bar{\alpha}_T = \prod\nolimits_{s=1}^{T} (1 - \beta_s). \label{eqn:diffusion_drv2}
\end{equation}
For inference, $\hat{\textbf{v}}_{joint_{E}}$ is sampled from a noisy initialization with a denoising network $\epsilon_\theta$, assisted by a conditioning vector $\textbf{c}$ via formula:
\small
\begin{equation}
\!\!\!\!\!\!\!\hat{\mathbf{v}}_{joint_{E}, t-1}\!=\!\frac{1}{\sqrt{\alpha_t}} \left(\!\hat{\mathbf{v}}_{joint_{E}, t}\!-\!\frac{1\!-\!\alpha_t}{\sqrt{1 \!-\!\bar{\alpha}_t}} \epsilon_\theta(\hat{\mathbf{v}}_{joint_{E},t}, t, \textbf{c})\! \right)\!,\!\label{eqn:diffusion_drv} 
\end{equation}
where $t \in [0,T]$ is the iterative denoising process, $\alpha_t = 1 - \beta_{t}$. Vector
$\textbf{c}$ has the same shape as $\textbf{v}_{joint_{E}}$, and is extracted from ${\hat{\textbf{x}}}$ by a DRV extractor $E_{C}$ that is forked from $E_{L}$ but has a modified input dimension to take only ${\hat{\textbf{x}}}$ as input.
Since our DRV only needs to provide complementary fidelity information instead of modeling entire image structures, a simple denoising network $\epsilon_\theta$ consisting of 4 ResMLP blocks with a 4-step sampling scheduler is used, largely reducing computation and memory requirements compared with conventional UNet-based SD denoising.
$\hat{\textbf{v}}_{joint_{E}}$ is then embedded into the Restormer~\cite{zamir2022restormer} for latent enhancement via additional cross-attention blocks inserted into the inner transformer blocks where $\hat{\textbf{v}}_{joint_{E}}$ serves as $key$ and $value$. Finally, the enhanced latent is fed into the LIC decoder to generate the final output image ${\hat{\textbf{x}}}_{lic}$ from the LIC stream, serving as the interactive baseline with the codebook-based generative stream for fidelity infusion.
\begin{figure}[hbp]
\centering
    \includegraphics[width=\linewidth]
    {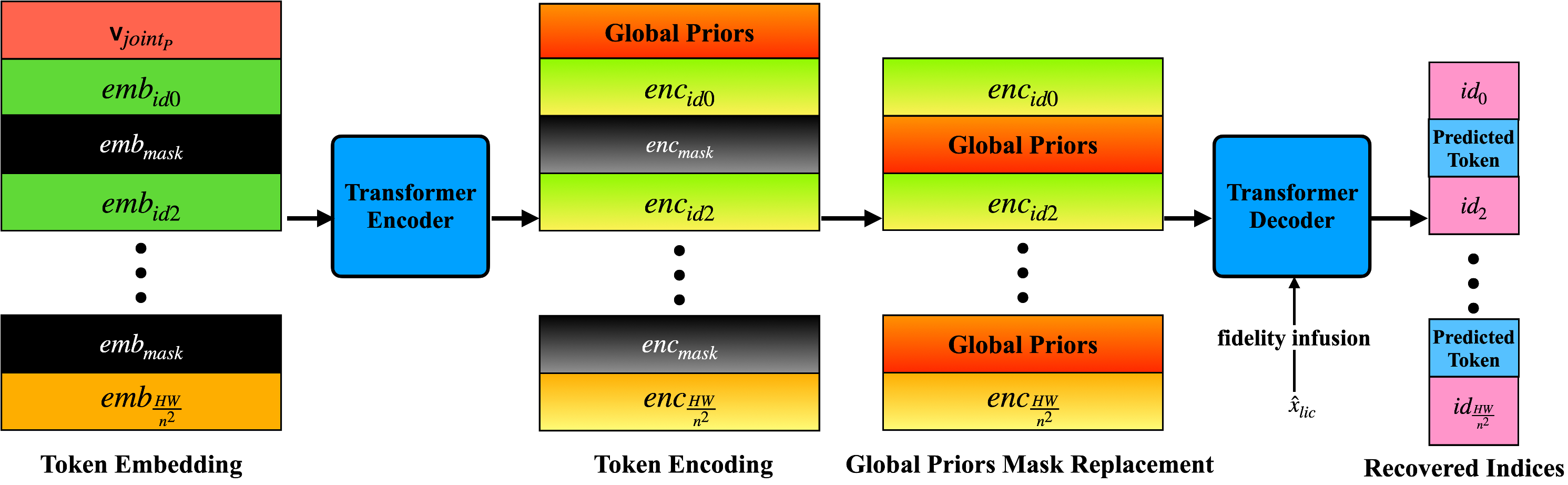}
    \caption{DRV $\hat{\textbf{v}}_{joint_P}$ embedding process in $\textbf{T}$.}
    \label{fig:transformer_detail}
\end{figure}
\subsection{The Generative Stream}
In general, this stream encodes image $\textbf{x}$ into a discrete indices map via a codebook-based representation. First, a VQ-encoder $\textbf{E}_{vq}$ encodes $\textbf{x}$ into a latent representation $\textbf{y}_{vq} \in \mathbb{R}^{\frac{H}{n}\times\frac{W}{n}\times C_{vq}}$ with the downsampling factor of $n$. Then $\textbf{y}_{vq}$ is further mapped into an indices map $\textbf{d}_{vq} \in \mathbb{R}^{\frac{H}{n} \times \frac{W}{n}}$ via a learned codebook with $K$ codewords $\textbf{C} = \{ c_{k} \in \mathbb{R}^{1\times C_{vq}}\}_{k=0}^{K}$. Each vector $y_{ij} \in \mathbb{R}^{1 \times C_{vq}}$ ($i \in [0, \frac{H}{n}], j \in [0, \frac{W}{n}]$) is mapped to the nearest codeword $c_{k} \in \textbf{C}$ by:\vspace{-1em}

\begin{equation}
    {\arg\min}_{k} \ \ \| c_{k} - y_{ij} \|.
\end{equation}
Transmitting whole $\textbf{d}_{vq}$ can be too costly for ultra-low-bitrate scenarios. For example for $K$=1024 and $n$=16, the bpp is about $10/256 \!\!\approx\!\!0.04$, which is close to the upper bound of ultra-low bitrate range ($<$0.05). Thus, compression-friendly binary masks have been used~\cite{lu2024hybridflow} to transmit only a masked portion of $\textbf{d}_{vq}$: $\textbf{d}_{masked} = \textbf{m} \circ \textbf{d}_{vq}$.

Then in the decoding stage, an indices map $\hat{\textbf{d}}_{vq}$ is recovered from $\textbf{d}_{masked}$ by masked index prediction through a token-based encoder-decoder transformer $\textbf{T}$. A codebook-based latent ${\hat{\textbf{y}}}_{vq}$ can be retrieved from the shared learned codebook $\textbf{C}$ based on indices ${\hat{\textbf{d}}}_{vq}$, which is used to reconstruct the output image by a VQ-decoder $\textbf{D}_{vq}$.
In our generative stream, the baseline ${\hat{\textbf{x}}}_{lic}$ from the LIC stream is used in two different ways to provide fidelity information: for indices map prediction, and for conditioned pixel decoding with VQ-latent correction.

\subsubsection{LIC-assisted indices map prediction}
The token-based encoder-decoder transformer $\textbf{T}$ takes as input a token-embedding $\textbf{emb}_{P}$ with length of $\frac{HW}{n^2}\!+\!1$ by $\{emb_{token_0}, \\
emb_{id_0}, emb_{mask}, emb_{id_3}, \cdots, emb_{id_{\frac{HW}{n^2}}}\}$, where $mask$ is the masked index to predict and $emb_{mask}$ is the embedding vector of $mask$; $id_{i}$ is the unmasked ground-truth index and $emb_{id_i}$ is its embedding vector; and $token_0$ is a  class token originally designed for class-based generation and $emb_{token_0}$ is the embedding vector of $token_0$. HybridFlow~\cite{lu2024hybridflow} ignores $token_0$ by using a fake class label without actual meaning. However, as shown in MAGE~\cite{li2023mage}, instead of fake `dummy' embedding, global priors can be fused into masked locations for improved prediction. Therefore, we fuse information from ${\hat{\textbf{x}}}$ into $\textbf{emb}_{P}$, which serves as image-aware global embedding for better prediction. 

Specifically, similar to the case of the DRV extractors $E_L$ in the LIC stream, a DRV extractor $E_{P}$ extracts $\textbf{v}_{joint_{P}}$ from the concatenation of $\textbf{x}$ and $\hat{\textbf{x}}$.
$\textbf{v}_{joint_{P}}$ is then fused into the token-embedding 
\begin{equation}
    \textbf{emb}_P = \{\textbf{v}_{joint_{P}}, emb_{id_0}, emb_{mask}, \cdots, emb_{id_{\frac{HW}{n^2}}}\}.
\end{equation}
Based on $\textbf{emb}_{P}$, the transformer encoder computes a token encoding as 
\begin{equation}
    \textbf{enc}_{P} = \{ enc_{\mathbf{v}}, enc_{id_0}, enc_{mask}, \cdots, enc_{id_{\frac{HW}{n^2}}} \},
\end{equation}
where $enc_{\mathbf{v}}$, $enc_{id_i}$, and $enc_{mask}$ correspond to the encoded DRV, encoded $id_i$, and encoded $mask$ respectively. To fully utilize the encoded global information from DRV, we replace all $enc_{mask}$ with $enc_{\mathbf{v}}$ in $\textbf{enc}_{P}$ before feeding it to the transformer decoder, so that all masked locations have global priors for better token prediction.

Fig.~\ref{fig:transformer_detail} illustrates the detailed process of DRV $\textbf{v}_{joint_P}$-guided masked token prediction using $\textbf{T}$. Note that the 2D VQ indices map $\textbf{d} \in \mathbb{R}^{\frac{H}{m} \times \frac{W}{m}}$ is first flattened into a sequence before being fed into $\textbf{T}$ in transformer.
Since LIC-assisted Indices Map Prediction is a form of conditional generation and high-level global priors are already fused into the masked locations via the $\textbf{v}{joint_P}$ embedding, we find it beneficial to further constrain the prediction output to more closely align with the ground truth. This is achieved by integrating lower-level fidelity features from the enhanced LIC output $\hat{\textbf{x}}_{lic}$ into the transformer decoder. Specifically, inspired by the “Prediction Assistance” design in HybridFlow~\cite{lu2024hybridflow}, we use a forked VQ Encoder $\textbf{E}_{P}$, with the output channel $\textbf{C}_{vq}$ modified to match the hidden dimension $h_{dim}$ of the transformer, to extract a lower-level feature map $\textbf{fm}_{P} \in \mathbb{R}^{\frac{H}{n} \times \frac{W}{n} \times h_{dim}}$ from $\hat{\textbf{x}}_{lic}$. The feature map $\textbf{fm}_{P}$ is then flattened to the shape $\frac{HW}{n^2} \times h_{dim}$ and fed into the cross-attention modules of the transformer decoder, serving as the $key$ and $value$.

To avoid transmitting $\textbf{v}_{joint_{P}}$ and save bits, we use a diffusion process, similar to the approach of using ground-truth infused DRV $\textbf{v}_{gt}$ for enhancing the LIC stream. We sample a $\hat{\textbf{v}}_{joint_{P}}$ in decoder from a noisy initialization, conditioned on a vector $\textbf{c}_{P}$ extracted from $\hat{\textbf{x}}_{lic}$ by following Eq.(\ref{eqn:diffusion_drv1}$\sim$\ref{eqn:diffusion_drv}). The denoising module for generating $\hat{\textbf{v}}_{joint_{P}}$ has the same simple network structure and sampling scheduler as the denoising module for generating $\hat{\textbf{v}}_{joint_{E}}$, and the network for extracting $\textbf{c}_{P}$ from $\hat{\textbf{x}}$ share the same architecture with network $E_C$ for extracting $\textbf{c}$ from $\hat{\textbf{x}}$ in the LIC stream.

\subsubsection{LIC-assisted conditioned pixel decoding} The output image from the LIC stream ${\hat{\textbf{x}}}_{lic}$ provides important fidelity information to the pixel decoding process for reconstructing the final image. We propose an $S$-channel VQ-correction module guided by $\hat{\textbf{x}}_{lic}$ to mitigate the VQ loss caused by inaccurate codebook-entry mapping and introduce an assistive decoder $\textbf{D}_{a}$ to infuse the fidelity information to the VQ-Decoder $\textbf{D}_{vq}$. 

\begin{figure}[hbp]
\centering
    \includegraphics[width=\linewidth]
    {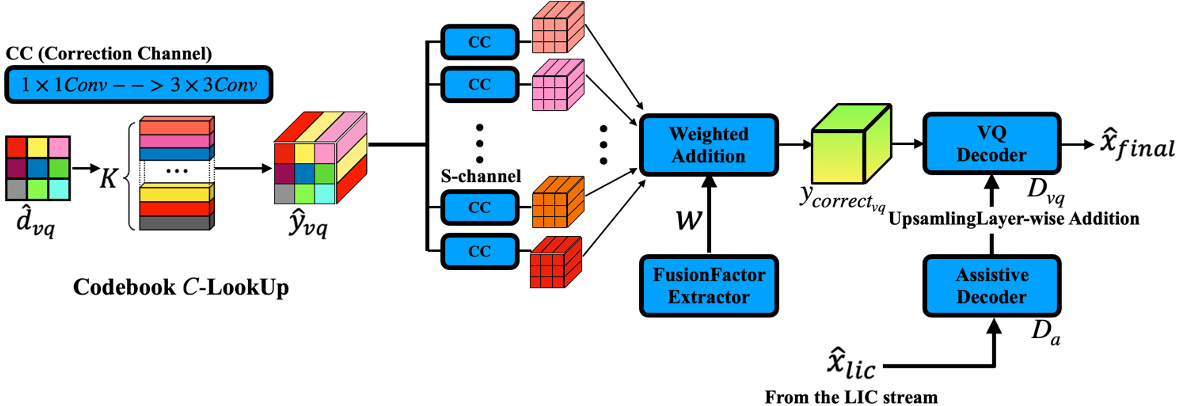}
    \caption{VQ Correction Module for dual-stream merging.}
    \label{fig:vq_detail}
\end{figure}

The detailed structure of the VQ-Correction module is shown in Fig.~\ref{fig:vq_detail}, where $\hat{\textbf{y}}_{vq}$ is retrieved in decoder from the codebook using the recovered indices $\hat{\textbf{d}}_{vq}$. It is then fed into $S$-parallel channels, each comprising a $3\!\times\!3$ and $1\!\times\!1$ conv kernel, to generate $S$ derived latents $\textbf{ys}_{1},\ldots,\textbf{ys}_{S}$. Each $\textbf{ys}_{i}\in \mathbb{R}^{\frac{H}{n} \times \frac{W}{n} \times C_{vq}}$ has the same shape as $\hat{\textbf{y}}_{vq}$. Then they are weighted to give a final corrected VQ latent: 
\begin{equation}
\textbf{y}_{correct_{vq}}\!\in\!\mathbb{R}^{\frac{H}{n} \times \frac{W}{n} \times C_{vq}} \!= \!\sum\nolimits_{i=1}^{S} \!\left( \textbf{ys}_{i,\frac{H}{n},\frac{W}{n},C_{vq}} \cdot \textbf{w}_{i,\frac{H}{n},\frac{W}{n}} \right)
\end{equation}
Combining weights $\textbf{w}\!\in\!\mathbb{R}^{S\times\frac{H}{n}\times\frac{W}{n}}$ are extracted from $\hat{\textbf{x}}_{lic}$ by a weight extractor comprising several residual swin transformer blocks (RSTB). $\textbf{w}$ carries input-adaptive information from $\hat{\textbf{x}}_{lic}$ to reduce indices mapping loss in $\textbf{y}_{correct_{vq}}$. 

The assistive decoder $\textbf{D}_{a}$ has the same structure as the VQ-Decoder $\textbf{D}_{vq}$.  $\textbf{D}_{a}$ takes in $\textbf{y}_{correct_{vq}}$ and the feature output after each upsampling layer is element-wisely added to the corresponding layer of $\textbf{D}_{vq}$ via connection links. $\textbf{D}_{vq}$ decodes $\textbf{y}_{correct_{vq}}$ together with additional information from $\textbf{D}_{a}$ to reconstruct the final output image ${\hat{\textbf{x}}}_{final}$. Note that we remove $Softmax()$ 
 to improve module performance during the training phase. Thus, the direct interaction between the $\textbf{w}$ extracted from the enhanced LIC output $\hat{\textbf{x}}_{lic}$ facilitates a more seamless integration of the enhanced LIC feature space into the VQ-based generative feature space. 
It is worth mentioning that our method has the identical bit consumption as the dual-stream HybridFlow~\cite{lu2024hybridflow}, since both the ground-truth DRV $\textbf{v}_{gt}$ for LIC stream enhancement and latent $\textbf{v}_{join_{P}}$ for improved indices prediction are reconstructed in the decoder. In other words, by learning diffusion priors, we enhance dual-stream performance without additional transmission overhead. 

\subsection{Training Pipeline} 
\begin{figure*}[htbp]
\centering \includegraphics[width=\linewidth]
    {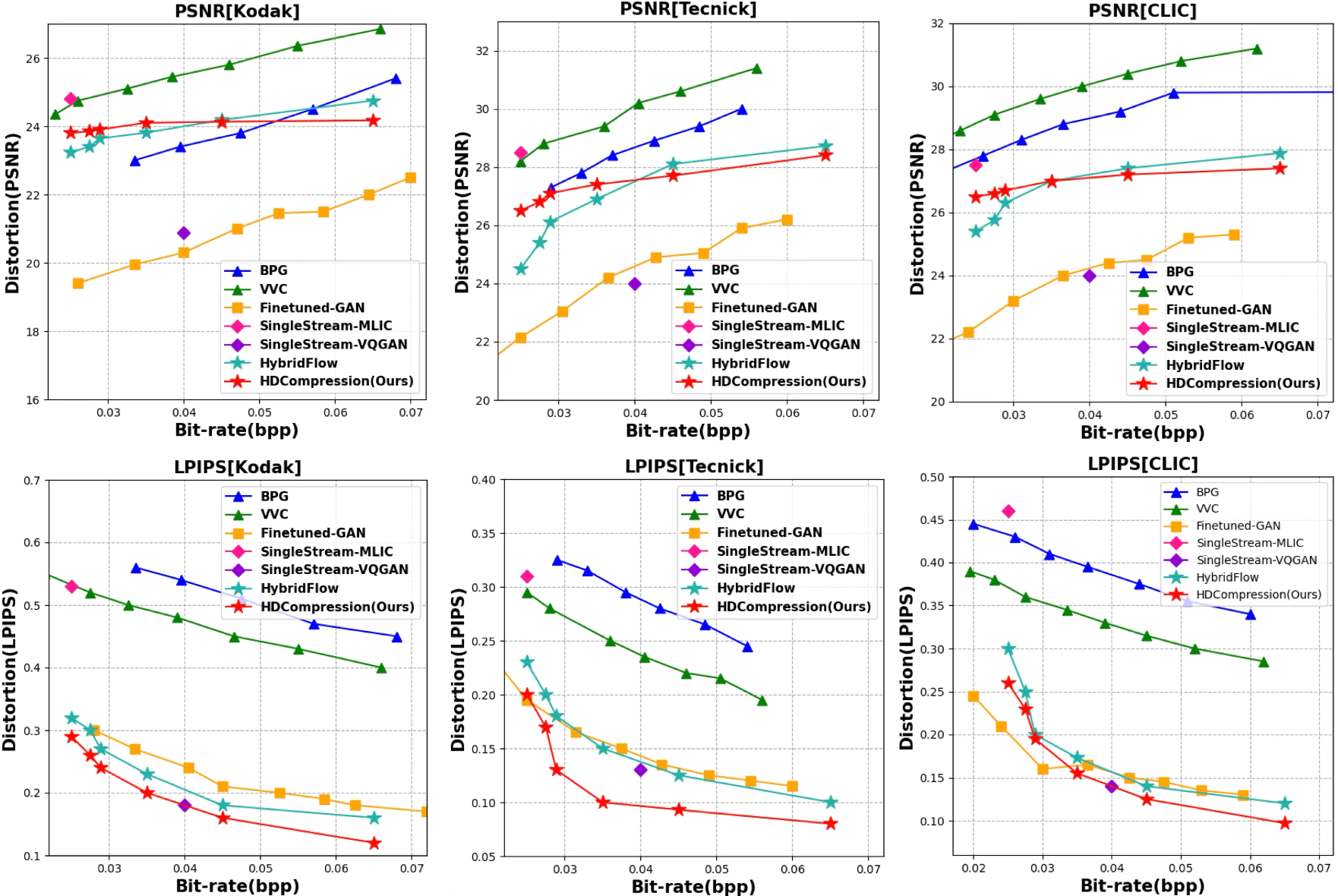}
    \caption{Quantitative metrics on Kodak, Tecnick and CLIC2020 test set. \textbf{PSNR} the higher the better. \textbf{LPIPS} the lower the better.}
    \label{fig:kodak_data}
\end{figure*}
Our entire framework is trained through multiple stages to balance training effectiveness and efficiency. 

\subsubsection{Basic flow pre-training} 
The LIC stream uses the pre-trained LIC encoder from MLIC. The VQ-Encoder $\textbf{E}_{vq}$ and the learned visual codebook $\textbf{C}$ use the pre-trained VQGAN model~\cite{esser2021taming}. These pre-trained components are designed to either pursue high-fidelity reconstruction or high-quality reconstruction, ensuring the baseline performance of our dual-stream system.

\subsubsection{DRV-based enhancement module training} 
The extractor $\textbf{E}_{L}$ for DRV $\textbf{v}_{joint_{E}}$ and the enhancement module are jointly trained based on the difference between the enhanced $\hat{\textbf{x}}_{lic}$ and the ground-truth $\textbf{x}$, with image loss:
\begin{equation}
    \label{eq: imageloss}
        L_{E} \!=\!w_{1} \!*\!L_{1} \!+\!w_{2}\!*\!L_{P}
        + w_{3}\!*\!L_{G},
\end{equation}
where $L_{1}$, $L_{P}$, and $L_{G}$ are L$_{1}$ pixel loss, perceptual loss via AlexNet, and UNet-based pixel-wise discriminator GAN loss, between ${\hat{\textbf{x}}}_{lic}$ and $\textbf{x}$, weighted by $w_1$, $w_2$, and $w_3$, respectively.
\subsubsection{DRV-based transformer predictor training} 
A binary mask $\textbf{m}_{b}$ is randomly selected among pre-fix mask schedulers and applied to the indices map $\textbf{d}_{vq}$ generated by pre-trained VQ-Encoder $\mathbf{E}_{vq}$  and codebook $\mathbf{C}$. The extractor $\textbf{E}_{P}$ for DRV $\textbf{v}_{joint_{P}}$ and the encoder-decoder transformer $\textbf{T}$ are trained together to predict the masked tokens assisted by $\textbf{v}_{joint_{P}}$, with loss:
\begin{equation}
\label{eq:tokenlogitloss}
L_{T}= -E(\sum \mathrm{log} p(m_{i}|d_{r}, \textbf{v}_{joint_{P}})),
\end{equation}
where $m_{i}$ is the predicted masked tokens and $d_{r}$ is the remaining unmasked indices.
\subsubsection{VQ correction module training} We keep the pre-trained VQ-encoder $\mathbf{E}_{vq}$ and codebook $\mathbf{C}$ frozen and train the VQ-decoder $\mathbf{D}_{vq}$ together with the VQ-correction module and the assistive decoder $\mathbf{D}_a$, based on the difference between the final output $\hat{\textbf{x}}_{final}$ and the ground-truth $\textbf{x}$ with a similar loss function as Eq.~\ref{eq: imageloss} to ensure more stable performance.

\subsubsection{DRV-diffusion module training} The two DRV-diffusion modules, one for enhancing the LIC stream and another for indices map prediction, where each learns an independent 4-step DRV-based diffusion process.
We optimize the mean squared error (MSE) between denoised DRV vector ($\hat{\textbf{v}}_{joint_{E}}$ or $\hat{\textbf{v}}_{joint_{P}}$) and its ground-truth ($\textbf{v}_{joint_{E}}$ or $\textbf{v}_{joint_{P}}$), using pre-trained joint-DRV extractors ($\textbf{E}_{L}$ or $\textbf{E}_{P}$). Instead of focusing on specific steps, our approach minimizes the cumulative loss after the full denoising process, ensuring better DRV reconstruction quality.
\begin{figure*}[htbp]
\centering    
\includegraphics[width=\linewidth]   {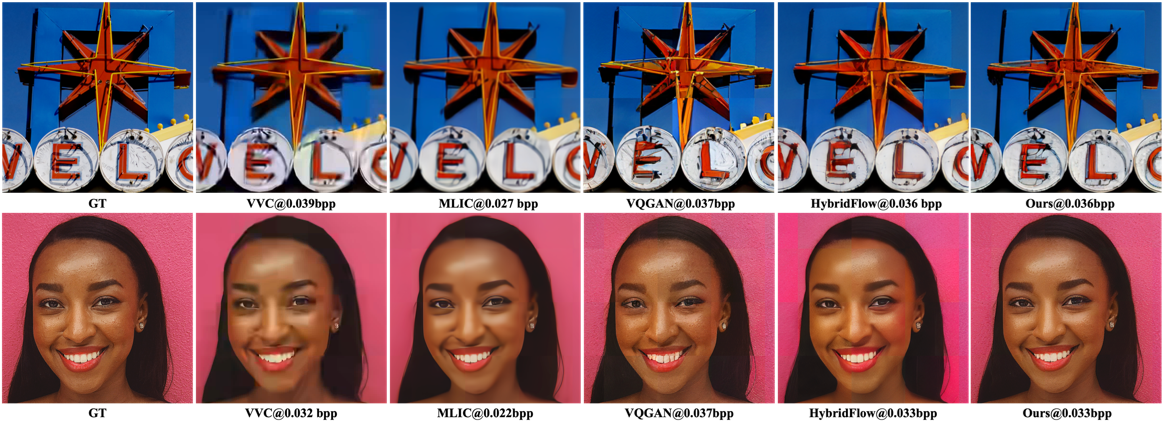}
    \caption{Qualitative Comparison of our method to the baselines. "1\_4" mask strategy ($75\%$ mask ratio on $\textbf{d}_{vq}$) is utilized to maintain around 0.035 bpp within the similar range of the compared baselines. Zoom in for better visualization.}   
    \label{fig:qualitative_compare}
\end{figure*}

\section{Experiments}
\label{sec:exp}
\begin{figure*}[htbp]
\centering
\includegraphics[width=1\linewidth]
    {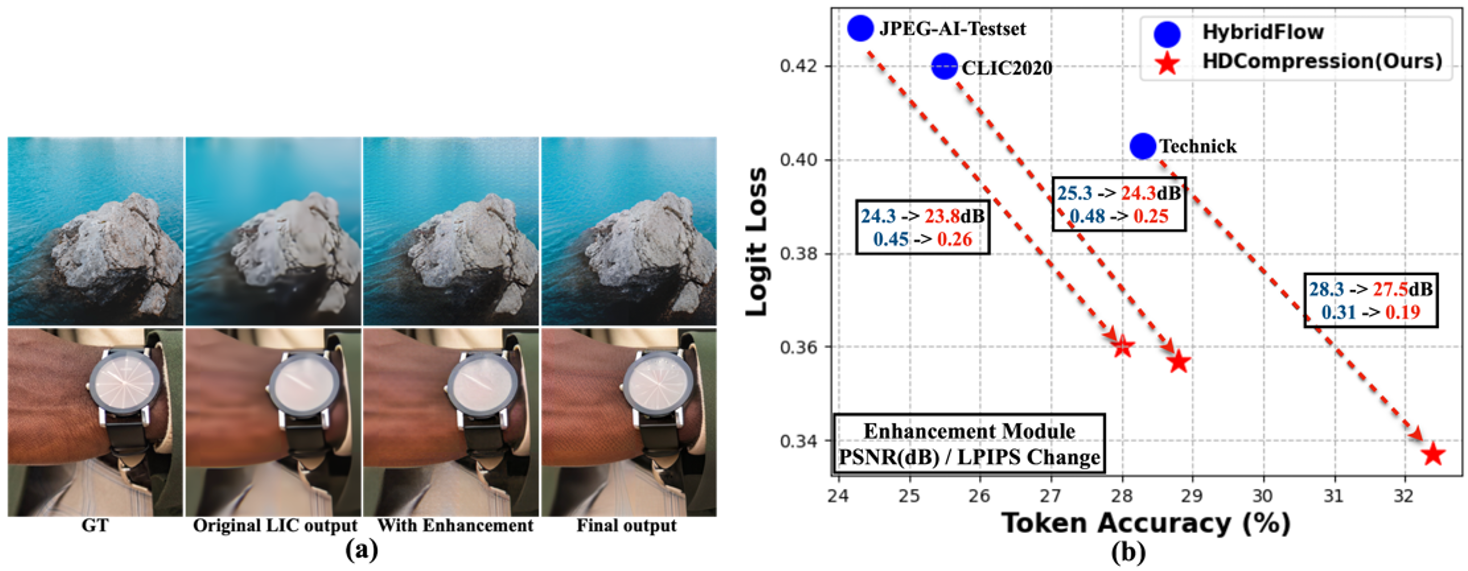}
    \caption{Impact of hybrid-diffusion modules. \textbf{(a)}: Visual improvement of DRV-based enhancement module over base LIC and effect of VQ-correction merging. \textbf{(b)}: Increase of the token-prediction accuracy via DRV-based mask predictor and quantitative improvements from DRV-based enhancement module across various datasets.}
  \label{fig:visualization_self}
\end{figure*}
\noindent \textbf{Datasets.} Our HDCompression model is trained on ImageNet-1k~\cite{imagenet}, with 1000 categories. In alignment with HybridFlow~\cite{lu2024hybridflow}, performance is evaluated over three benchmarks: Kodak~\cite{kodak}, CLIC 2020 test set~\cite{CLIC2020} and Tecnick dataset~\cite{tecknick}. 

\noindent \textbf{Model configurations.}
Training images are cropped into $256\!\times\!256$ patches. To achieve approximately 0.025 bpp using pre-trained MLIC models~\cite{mlic2023} as the Base LIC, which typically provide a minimum of 0.1 bpp, input patches for the LIC stream are further downsized to $128 \times 128$ via bilinear interpolation. Both streams have the same downsampling factor $m = n = 16$. The lightweight diffusion modules use a 4-step DDPM scheduler. We unify the loss weights as follows: $w_{1} = 1.2$ for $L_{1}$ pixel loss, $w_{2} = 0.8$ for AlexNet-based LPIPS perceptual loss, and $w_{3} = 0.12$ for UNet-based pixel-wise discriminator GAN loss. 

\noindent \textbf{Compared baselines.} 
We compare with several representative image compression methods: 1) Traditional hand-crafted VVC \& BPG; 2) MLIC~\cite{mlic2023}, a single-streamed conventional LIC method; 3) VQGAN~\cite{esser2021taming} \& Fine-tuned VQGAN~\cite{VQPeking}, single-streamed VQ-codebook-based methods; and 4) HybridFlow~\cite{lu2024hybridflow}, a dual-stream framework that straightforwardly combines pre-trained LIC with VQ-codebook-based stream. To obtain ultra-low bitrates, we set $\textbf{QP}$ ranging in $[45,51]$ for VVC \& BPG, and downsize the input image (1/2 width and height) for MLIC.

\noindent \textbf{Evaluation metrics.} 
We evaluate commonly used PSNR and LPIPS~\cite{LPIPS}. PSNR measures pixel-level distortion and LPIPS assesses the visual quality. 
\subsection{Quantitative Results}
HDCompression has the same bpp as HybridFlow, which operates over $[0.025, 0.065]$ bpp range, spanning from fully masked indices map (lowest quality) to unmasked indices map (highest quality). As shown in Fig.~\ref{fig:kodak_data}, traditional handcrafted VVC \& BPG and conventional MLIC outperform codebook-based methods over PSNR, due to their learning target of minimizing pixel-level distortions. However, codebook-based methods perform significantly better for perceptual LPIPS. The distortion-driven focus gives artificially inflated PSNR, which omits image details and prefers overly smoothed regions. In contrast, single-stream codebook-based methods emphasize LPIPS, neglecting fidelity to the original image, resulting in $>$4 dB PSNR drop compared with traditional methods at the same bpp.

HybridFlow attempts to balance PSNR and LPIPS, which increases PSNR by about 3 dB compared with single-stream codebook-based methods while offering better LPIPS. Our HDCompression further improves LPIPS through diffusion models, providing visually more pleasing reconstruction, and meanwhile maintaining a stable PSNR curve. With the increase of bpp, the generative stream offers more ground-truth information to compensate for the conventional LIC stream and retain only important general fidelity information.
Overall, our HDCompression achieves approximately 26\% LPIPS improvement compared to HybridFlow while preserving the same level of PSNR, providing a better balance between fidelity and perceptual quality under ultra-low-bitrate conditions.
\subsection{Qualitative results} 
As shown in Fig.~\ref{fig:qualitative_compare}, HDCompression makes more realistic and sharper image reconstructions compared to other baseline methods. Specifically, VVC and MLIC suffer from significant blurs for heavy rounding quantization. To maintain pixel-wise fidelity, they generate highly smoothed color blocks that are perceptually unpleasant. The single-stream VQGAN fabricates inauthentic details in sensitive regions, \textit{e.g.}, the star-shaped details in the first row.  HybridFlow partially addresses these problems but still suffers from excessive smoothing and detail loss due to the direct utilization of low-quality LIC as assistive information. Our HDCompression effectively resolves such issues with more effective dual-stream fusion, significantly surpassing the reconstruction quality of HybridFlow. Additionally, HDCompression mitigates boundary effects compared to VQGAN and HybridFlow, making block-wise fragmentation less noticeable.
\subsection{Ablation Study on Hybrid-Diffusion Modules}
We conduct an ablation study against the dual-stream HybridFlow to investigate the impact of our diffusion modules.

\subsubsection{DRV-based enhancement for the LIC stream}
We compare our LIC stream output ${\hat{\textbf{x}}}_{lic}$ of incorporating the enhancement module against the original output ${\hat{\textbf{x}}}$ of the pre-trained MLIC. As illustrated in Fig.~\ref{fig:visualization_self} (a), the enhancement module improves the quality of the original LIC output ${\hat{\textbf{x}}}$, particularly by reducing blurs. The enhancement module significantly enhances LPIPS of ${\hat{\textbf{x}}}$ while maintaining almost the same PSNR across various datasets in Fig.~\ref{fig:visualization_self} (b).

\subsubsection{DRV-based transformer for mask prediction}
We compare the logit-wise token prediction loss (Eq.~\ref{eq:tokenlogitloss}) of the DRV-based transformer mask predictor against the naive transformer mask predictor in HybridFlow without using DRV. As shown in Fig.~\ref{fig:visualization_self} (b) where the indices map is masked by "1\_4" masking schedule (75\% masking ratio), the prediction loss is dropped by 18.5\% on average by using DRV in the transformer encoder, leading to 15\% accuracy improvement in token prediction on average. Thus, more ground-truth indices are recovered to provide more specific details to the generative stream that might be neglected by the LIC stream.

\subsubsection{VQ correction for dual-stream merging} Even though the enhancement module effectively improves the quality of the LIC stream as stated above, the poor quality of the original LIC output still results in the loss of details, systematic noise artifacts, \textit{etc.}, the missing details of the watch and the blocky sea surface in Fig.~\ref{fig:visualization_self} (a). It is difficult for the enhancement module to recover the significant information loss from the rounding quantization at ultra-low bitrates. When the generative VQ-based information is merged with the enhanced LIC stream, details are further appended and the artifacts are largely removed in the final output. By the high-frequency-friendly generative information infused from the generative stream, the merging process sharpens the enhanced LIC output, making it more visually pleasing to the human eye.

\section{Conclusion}
\label{sec:conclusion}
In this paper, we have proposed HDCompression, a hybrid dual-stream framework that integrates the lightweight DRV-diffusion modules for ultra-low-bitrate image compression. The DRV-guided enhancement module effectively improves the quality of the fidelity information provided by the LIC stream. The DRV-guided token mask predictor increases the token prediction accuracy. The DRVs are reconstructed in the decoder via a conditioned diffusion process to avoid transmission overhead. The VQ-correction module infuses fidelity information from the enhanced LIC stream into the VQ-codebook-based generative stream for improved faithfulness in reconstruction.  Experiments have demonstrated significantly improved perceptual quality (LPIPS) with the same level of fidelity (PSNR) compared to previous dual-stream methods.

%
%
%
\bibliographystyle{splncs04}
\bibliography{refs}

\end{document}